\documentclass[10pt]{article} 
\usepackage[preprint]{tmlr}

\usepackage{amsmath,amsfonts,bm}

\def\eqref#1{equation~\ref{#1}}

\def\1{\bm{1}}

\DeclareMathAlphabet{\mathsfit}{\encodingdefault}{\sfdefault}{m}{sl}
\SetMathAlphabet{\mathsfit}{bold}{\encodingdefault}{\sfdefault}{bx}{n}

\usepackage{hyperref}
\usepackage{url}
\usepackage{geometry}
\newgeometry{vmargin={25mm}, hmargin={35mm,35mm}}
\usepackage{graphicx}%
\usepackage{multirow}%
\usepackage{amsmath,amssymb,amsfonts}%
\usepackage{amsthm}%
\usepackage{bbold}%
\usepackage{mathrsfs}%
\usepackage[title]{appendix}%
\usepackage{xcolor}%
\usepackage{textcomp}%
\usepackage{manyfoot}%
\usepackage{booktabs}%
\usepackage{algorithm}%
\usepackage{algorithmicx}%
\usepackage{algpseudocode}%
\usepackage{listings}%
\usepackage{makecell}%
\usepackage{float}%
\usepackage{booktabs}
\usepackage{lmodern}
\usepackage{rotating}

\title{Multimodal Prescriptive Deep Learning}

\author{\name Dimitris Bertsimas \email dbertsim@mit.edu \\
      \addr {Sloan School of Management and Operations Research Center, Massachusetts Institute of Technology, Cambridge, MA}
      \AND
      \name Lisa Everest \email leverest@mit.edu \\
      \addr {Operations Research Center, Massachusetts Institute of Technology, Cambridge, MA}
      \AND
      \name Vasiliki Stoumpou \email vasstou@mit.edu\\
      \addr {Operations Research Center, Massachusetts Institute of Technology, Cambridge, MA}}

\def\month{01}  
\def\year{2025} 
\def\openreview{\url{https://openreview.net/forum?id=XXXX}} 

\begin{document}

\maketitle

\begin{abstract}
We introduce a multimodal deep learning framework, Prescriptive Neural Networks (PNNs), that combines ideas from optimization and machine learning, and is, to the best of our knowledge, the first prescriptive method to handle multimodal data. The PNN is a feedforward neural network trained on embeddings to output an outcome-optimizing prescription. In two real-world multimodal datasets, we demonstrate that PNNs prescribe treatments that are able to significantly improve estimated outcomes in transcatheter aortic valve replacement (TAVR) procedures by reducing estimated postoperative complication rates by 32\% and in liver trauma injuries by reducing estimated mortality rates by over 40\%. In four real-world, unimodal tabular datasets, we demonstrate that PNNs outperform or perform comparably to other well-known, state-of-the-art prescriptive models; importantly, on tabular datasets, we also recover interpretability through knowledge distillation, fitting interpretable Optimal Classification Tree models onto the PNN prescriptions as classification targets, which is critical for many real-world applications. Finally, we demonstrate that our multimodal PNN models achieve stability across randomized data splits comparable to other prescriptive methods and produce realistic prescriptions across the different datasets. 
\end{abstract}

\section{Introduction}\label{sec:introduction}

Today's society provides an increasing availability of large quantities of data, particularly multimodal data consisting of structured and unstructured elements. As a result, developing systematic and personalized decision-making methods that can leverage such multimodal data becomes more and more critical, and the benefits of data-driven methods become more and more visible. For example, medical professionals could systematically and optimally treat patients based on individual characteristics, clinical notes, and medical scans \citep{haim}. Companies in technology and digital advertising would be able to increase customer impact by customizing content and advertisements according to user-specific data. In the retail industry, such personalized models would allow companies to dynamically price goods and services based on the user or environment for increased revenue.

Much of the current work in machine learning and deep learning focuses on improving the accuracy of output prediction. We find that deep learning has tremendous and underutilized potential in the area of decision-making. This paper combines ideas from machine learning and optimization to move from prediction to prescription, with the ability to leverage multimodal data. We introduce a novel, multimodal, deep learning framework that we call a Prescriptive Neural Network (PNN). In this paper, we demonstrate how our models handle complex data structures and how effective they are in both multimodal and unimodal real-world applications. Through these applications, we show that our PNN models are flexible with different treatment scenarios that cover all real-life application settings. Our models also give stable and realistic results, comparable to existing prescriptive methods, and provide the user with more control over the resulting prescriptions. On tabular datasets in particular, we are able to recover interpretability by applying knowledge distillation and fitting interpretable Optimal Classification Tree models \citep{optimal-classification-trees,MLbook} on the PNNs' prescriptions as classification targets; we find that these Mirrored OCTs perform comparably to their PNN counterparts, meaning that interpretability comes with little cost to performance. 

\subsection{Related Literature}\label{subsec:related-literature}
Previous literature in data-driven personalized decision-making includes the Regress \& Compare framework, tree-based methods, and causal methods. 

\textbf{Regress \& Compare.} 
Regress \& Compare is a black-box methodology where a regression model is trained to predict the outcome under each treatment. The set of features used for training is the augmented feature data combined with the historical treatment given. Given an input observation and possible treatment options, the model is then used to select the treatment with the lowest (highest) outcome for a minimization (maximization) problem. 

There are many applications of the Regress \& Compare methodology for prediction -- examples include energy economics \citep{electricity-markets} and multidrug-resistant tuberculosis \citep{tuberculosis-causal-inference}. Other works use the Regress \& Compare framework to move from predictions to prescriptions. In particular, \citet{predictive-to-prescriptive-analytics} extends Regress \& Compare solutions for prescriptive problems and incorporates $k$-nearest neighbors regression ($k$NN \citet{knn}), local linear regression (LOESS \citet{loess}), classification and regression trees (CART \citet{cart}), and random forests (RF \citet{rf}). Although \citet{predictive-to-prescriptive-analytics} demonstrates that their methods are widely applicable and computationally tractable under mild conditions, we note that these are classical machine learning methods and do not take advantage of neural networks. 

More specific applications of Regress \& Compare for prescriptive problems include healthcare \citep{personalized-diabetes, heart-failure} and revenue management \citep{observational-data-pricing}. \citet{personalized-diabetes} considers personalized diabetes treatment, while \citet{heart-failure} combines prediction and decision-making to allocate interventions for post-discharge patients that were admitted due to heart failure. \citet{observational-data-pricing} considers the problem of optimal pricing, where they learn from historical observational data to optimize predicted revenue given price.

One possible limitation of the Regress \& Compare approach is that it is affected by the number of samples per treatment, since for each treatment, only the samples that received that treatment in real life are considered. Also, it does not address the potential treatment assignment bias present in the data; e.g. healthier patients tend to receive lighter treatment and to have better outcomes. This is discussed in more detail in Section \ref{subsec:counterfactual-estimation}. Furthermore, the black-box nature reduces its interpretability, which is important for many real-world applications.

\textbf{Tree-based methods.} \citet{recursive-partitioning} introduces Personalization Trees, which extend the Regress \& Compare method for the problem of choosing the treatment with the best causal effect from a finite number of discrete options. Kallus presents three different recursive-partitioning-based algorithms: a greedy Personalization Tree, a Personalization Forest that bags Personalization Trees, and a globally optimal Personalization Tree. As these are tree methods, we note that they preserve interpretability. 

\citet{optimal-prescriptive-trees} introduces Optimal Prescriptive Trees, which are similar to Personalization Trees but combine the counterfactual estimation and prescriptive learning tasks together in one training process and extend the framework of Optimal Classification Trees from \citet{optimal-classification-trees,MLbook}. As a result, the trees are highly interpretable. \citet{optimal-policy-trees} further explores the optimal trees methodology and proposes Optimal Policy Trees, in which counterfactual estimation is performed separately from the prescriptive learning task. This allows for greater flexibility in discrete and continuous treatments, as well as better learning of the prescriptive task due to reduced complexity that results from the separation of the two training tasks. Like Optimal Prescriptive Trees, this method also preserves interpretability. These approaches, however, struggle with learning more complicated functional forms and are therefore limited to learning outcome functions that can be modeled by trees of relatively small depth.

\textbf{Causal methods.} This family of methods originates from the causal inference literature and includes both individual trees (causal trees) and their combinations (causal forests). \citet{athey} introduces causal trees, which employ a recursive partitioning approach of the feature space to split the data into groups with similar treatment effects. Causal forests extend causal trees and represent a prescriptive black-box method that builds on random forests, as introduced by \citet{causal-forests}. While random forests are constructed from decision trees, causal forests are composed of causal trees, which aim to maximize the difference in outcomes between two treatments at each node during tree growth. The resulting outcomes are interpreted relatively to one another. In the binary treatment case, since there are only two options (treatment or no treatment), one option will yield a positive effect (outcome) and the other a negative effect. If the goal is to minimize the outcome, the treatment option with the negative effect is prescribed.

Other models in the causal inference literature include causal boosting \citep{powers_qian_jung_schuler_shah_hastie_tibshirani_2018} and causal MARS \citep{powers_qian_jung_schuler_shah_hastie_tibshirani_2018}. However, the estimation of treatment effects, which is achieved by causal models, is not an explicit policy prescription, which is the goal of this work. 

Finally, another approach by \citet{Zhou_Athey_Wager_2023} takes inspiration from the causal inference literature and uses inverse propensity weight estimators to calculate the counterfactuals. Decision trees (both greedy and fully optimal) are then used for policy learning. Fully optimal trees, however, struggle with scalability, while the heuristic-based trees do not guarantee the best possible policy (optimality).

\textbf{Deep learning methods.}
Other deep learning approaches to the optimal prescription problem include \citet{other-pnn, asterios}. \citet{other-pnn} introduces prescriptive networks that are shallow neural networks to address the binary treatment regime, in which a treatment may or may not be given. Their networks are optimized by overestimating conditional average treatment effects (CATE), and they propose a method using mixed-integer programming (MIP) to implement their networks into commercial solvers. \citet{asterios} proposes a piecewise linear neural network model to output optimal prescriptions from a set of discrete treatments and show that their model partitions the input space into disjoint polyhedra, where all observations in the same partition are assigned the same treatment. \citet{bergman2022janos} proposes a solver that takes as input user-specified pretrained predictive models (including neural networks) and formulates optimization models directly over those predictive models to provide final prescriptions. 

We note that, like our PNN models, all of these works successfully combine ideas from optimization and machine learning. However, \citet{bergman2022janos} does not incorporate Deep Learning in the prescriptive part of the framework, but only to generate predictions. \citet{other-pnn, asterios} consider binary and discrete treatments respectively, while our work handles more treatment and outcome scenarios. Furthermore, our approach differs in the network's objective function used for training.

\subsection{Contributions}\label{subsec:contributions}

Our contributions are as follows:
\begin{enumerate}
    \item Combining machine learning and optimization, we propose a novel, multimodal, deep learning framework we call a Prescriptive Neural Network (PNN); to the best of our knowledge, our model is the first prescriptive method to handle multimodal data.

    \item In two real-world multimodal datasets, we demonstrate that PNNs prescribe treatments that are able to significantly improve estimated outcomes in transcatheter aortic valve replacement (TAVR) procedures by reducing estimated postoperative complication rates by 32\% and in liver trauma injuries by reducing estimated mortality rates by over 40\%. Additionally, PNNs either outperform or perform comparably to existing, state-of-the-art, prescriptive methods on four real-world unimodal (tabular) datasets that span all four treatment scenarios: diabetes management (multiple continuous treatments), groceries pricing (single continuous treatment), splenic injuries treatment (multiple discrete treatments), and REBOAs in blunt trauma patients (binary treatment). 

    \item On tabular datasets we recover interpretability through knowledge distillation; we train Optimal Classification Trees (OCT) \citep{optimal-classification-trees,MLbook} on the feature data but using the PNN prescriptions as target classes, similar to a binary or multiclass classification task. We call these Mirrored OCTs. Remarkably, the performance of the Mirrored OCTs is equally strong as that of the original PNNs, with a decrease in improvement of only 1.38\% on average across the tabular datasets; this implies that interpretability may be recovered with minimal cost to performance. 
    
    \item Finally, we demonstrate that our multimodal PNN models achieve stability across randomized data splits comparable to other prescriptive methods and produce realistic prescriptions across the different datasets.
    
\end{enumerate}

\section{Methods}\label{sec:methods}

In this section, we review the methodology of our PNNs. We first formally define the prescriptive problem we seek to solve (Section \ref{subsec:problem_definition}), and then we present the training process, which is divided into four main steps: embedding extraction (Section \ref{subsec:embedding-extract}), counterfactual estimation (Section \ref{subsec:counterfactual-estimation}), prescription policy learning (Section \ref{subsec:prescription-policy-learning}), and interpretability recovery (Section \ref{subsec:oct-pnn}). 

\subsection{Problem definition}\label{subsec:problem_definition}

Formally, we consider a prescription problem, which can be characterized by observational data in the form $\{(\boldsymbol{x}_i,y_i,t_i)\}_{i=1}^n$:
\begin{itemize}
    \item \textbf{Features} $\boldsymbol{x}_{i}\in\mathbb{R}^p$ is the $p$-dimensional feature data for the $i$-th observation.
    \item \textbf{Treatment} $t_i\in\mathcal{T}$ is the treatment applied historically to the $i$-th observation, where $\mathcal{T}$ is the set of all possible treatments. As treatments may be discrete or continuous, there are four possible treatment scenarios: binary (treatment or no treatment), multiple discrete (two or more treatment options), single continuous (one treatment option with continuous values), or multiple continuous (two or more treatment options, some or all taking on continuous values). 
    \item \textbf{Outcome} $y_{i}\in\mathbb{R}$ is the result observed after treatment $t_i \in \mathcal{T}$ has been applied to the $i$-th observation. 
\end{itemize}
Given this observational data, the aim is to develop an optimal prescriptive model that outputs a treatment $t\in\mathcal{T}$ that results in an optimal outcome $y$ for each input observation $\boldsymbol{x}$. 

\subsection{Embedding Extraction}\label{subsec:embedding-extract}
The first step in the model pipeline is to extract embeddings from the structured and unstructured data.

\subsubsection{Structured data}
We extract embeddings from structured feature data through traditional preprocessing techniques as described below, where the technique depends on whether the feature is numerical, categorical, or ordinal.

\begin{itemize}
    \item \textbf{Numerical features.} Numerical features are normalized to the interval [0,1] for counterfactual estimation and model training to increase stability and equal weighting of features. We note that since tree models are independent of data scale, we use the original feature values when training all tree models, which ensure interpretability in the tree splits.
    \item \textbf{Categorical features.} For categorical features, we use one-hot encodings to convert them to binary features, such that each category becomes a new indicator feature.
    \item \textbf{Ordinal features.} Ordinal features are categorical features whose values carry numerical information. Since these categories have a natural order to them, we can assign each category a number such as 1 to 5, where relative magnitude holds information. The feature value assigned to the number ``1'' conveys that that value is less than that of a feature value assigned the number ``4.'' These ordinal features are then treated as numerical features in our experiments.
    
\end{itemize}

\subsubsection{Unstructured data} \label{subsubsec: unstructured}
We extract embeddings from unstructured data using pretrained, deep learning models. By passing each observation's unstructured data through these pretrained models, we can obtain a vector representation of the unstructured datapoint. In particular, our experiments on medical data in Section \ref{sec:realworld-datasets} use Clinical Longformer \citep{clinical-longformer}, a long sequence transformer model trained via a sparse attention mechanism on domain-specific, large-scale clinical corpora; from this model, we obtain a 768-dimensional embedding vector for each observation's text data. 

We then leverage Principal Component Analysis (PCA) for dimensionality reduction purposes. Though optional, this step helps to improve stability, tractability, and performance in our PNN model, since the embeddings extracted from the pretrained models can be high-dimensional,  which increases the complexity of the problem when the dataset size is comparable to the embedding dimension. In particular, we reduce to a 32-dimensional representation for the clinical notes in our experiments on medical data, but the size can be adjusted, depending on the application and the dataset size.

While we specifically use ClinicalLongformer, any pretrained large language model (LLM) may be used to process unstructured text data. Additionally, any pretrained computer vision (CV) model may be used to process unstructured image data. This results in an embedding extraction step for unstructured data that is not only highly accessible, but also highly flexible.

To get the final multimodal embeddings, we concatenate the individual modalities' embeddings to obtain one large embedding vector.

\subsection{Counterfactual Estimation}\label{subsec:counterfactual-estimation}
The next step is counterfactual estimation. Because the prescriptive problem's dataset only contains historical observational data, the counterfactuals are unknown, e.g. the hypothetical outcomes $y(\boldsymbol{x}_i, t)$ for $t \neq t_{i}$, for each observation $\boldsymbol{x}_i$. We therefore perform a counterfactual estimation step \citep{doubly-robust-policy} that estimates the outcomes for each observation under every treatment. This produces a rewards matrix $\Gamma$, where $\Gamma_{i,t}$ is the estimated outcome of applying treatment $t$ to the $i^{\text{th}}$ observation. The estimation process is slightly different for discrete and continuous treatments.

\subsubsection{Counterfactual estimation of discrete treatments}\label{subsubsec:counterfactual-estimation-discrete}

We use two methods for counterfactual estimation of discrete treatments. The doubly-robust method is, however, preferred for almost all of the experiments in Section \ref{sec:realworld-datasets}, since it addresses the treatment assignment bias. The two methods are as follows:

\begin{enumerate}
    \item \textbf{Direct Method.} This method directly learns the outcome function $y_{t}(\boldsymbol{x})$ by training separate models, one for each treatment $t$. During training, each model uses only the subset of the observations that received treatment $t$. These models can be random forests or boosting methods and output an estimated outcome $\hat{y_{t}}(\boldsymbol{x})$ for when treatment $t$ is hypothetically applied to observation $\boldsymbol{x}$. 

    \item \textbf{Doubly robust estimation.} Because direct estimation is often prone to treatment assignment bias, the doubly-robust estimator attempts to mitigate this bias by re-weighting the estimated direct outcomes with propensity score probabilities. This reweighting is expressed in Equation (\ref{eq:doubly-robust}), which calculates the doubly-robust reward matrix $\Gamma$:
    \begin{equation}\label{eq:doubly-robust}
        \Gamma_{i,t} = \hat{y}_{i,t} + \mathbb{1}\{t_{i}=t\} \frac{1}{p_{i,t}} (y_i-\hat{y}_{i,t}), 
    \end{equation}
    where $\hat{y}_{i,t} = \hat{y}_t(\boldsymbol{x}_i)$ is the estimated outcome of sample $i$ under treatment $t$, $p_{i,t} =  \mathbb{P}[t_i=t]$ is the probability that treatment $t$ is assigned to observation $i$ in real life and $y_i$ is the actual outcome of observation $i$. 
\end{enumerate}
For binary outcomes, we use classifiers for the counterfactual estimation, while for continuous outcomes, we use regressors. 

\subsubsection{Counterfactual estimation of continuous treatments}\label{subsubsec:counterfactual-estimation-continuous}

For continuous treatments, we train a regression model to predict the outcome of the $i^{\text{th}}$ observation using as input the observational data $\boldsymbol{x}_{i}$ and continuous prescribed treatment doses $T_{i,t}$ for each treatment $t$. Then, by discretizing the continuous dose values and only considering a subset of them as valid treatments, we use the trained model to retrieve the estimated outcome for the $i^{\text{th}}$ observation under all valid treatment schemes. This is analogous to real-world treatment scenarios; when we handle continuous treatments, we always select a subset of the possible ones, since the real-world treatment options need to be finite.

\subsection{Prescription policy learning through feedforward neural networks}\label{subsec:prescription-policy-learning}

\begin{figure}[!ht]
  \begin{center}
    \caption{Architecture of a feedforward neural network (\citet{fnn-image}).}\label{fig:fnn}
  \includegraphics[width=0.6\textwidth]{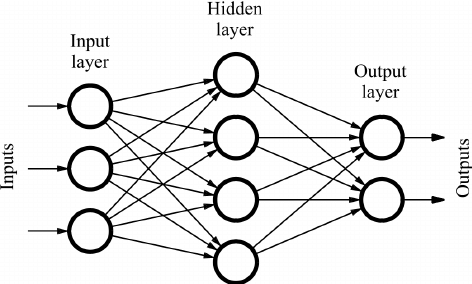}
  \end{center}
\end{figure}

A feedforward neural network consists of layers of interconnected neurons, which are computational units that are each characterized by weights $\{a_{j}\}_{j=1}^p$ (where $p$ denotes the number of neurons in the layer), a bias $b$, and a nonlinear activation function $h$. Given an input vector $\boldsymbol{x}\in\mathbb{R}^p$, the neuron calculates $y = \sum_{j=1}^{p}a_{j}x_{j} + b$, a weighted sum of the input's components, and passes the result $y$ through the activation function $h$ to produce the neuron's output $o=h(y)$ that serves as the input to the subsequent neuron. As observational inputs $\boldsymbol{x}$ pass through this network of interconnected neurons, this calculation is performed in a nested manner and the output of the final layer is computed, from which a pre-specified loss function is evaluated. Using backpropagation, the weights of the network are updated in order to minimize this loss function. The typical structure is presented in Figure \ref{fig:fnn}.

Without loss of generality, we assume our goal is to minimize outcomes in the prescriptive problem. The objective of our prescriptive neural network is to minimize total rewards for the prescriptions $\tau(\boldsymbol{x}_i)$ assigned by the network to each observation $\boldsymbol{x}_i$ in the dataset: 
\begin{equation}\label{eq:nn_obj_func}
    \min_{\tau(.)} \sum_{i=1}^{n} \sum_{t\in\mathcal{T}}\mathbb{1} \{\tau(\boldsymbol{x}_i) = t\}\cdot \Gamma_{i,t}, 
\end{equation}
proposed by \citet{optimal-policy-trees}. Because the indicator function is not differentiable, the backpropogation algorithm cannot handle Equation (\ref{eq:nn_obj_func}) exactly. We therefore ``soften'' the objective and leverage an approach analogous to that of multi-classification networks. The PNN assigns treatments probabilistically, such that its output layer consists of $|\mathcal{T}|$ neurons, one for each distinct treatment (as multi-classification networks have an output corresponding to each target class). We denote the output vector of the PNN as $\boldsymbol{z}\in\mathbb{R}^{|\mathcal{T}|}$ and apply a softmax activation function to these output neurons to obtain a probability distribution over the distinct treatments, such that $\mathbb{P}[\tau(\boldsymbol{x}_i) = t] = \sigma_t(\boldsymbol{z})$, where $\sigma(\cdot)$ denotes the softmax function and is defined explicitly below:  
\begin{equation}
    \sigma_t(\boldsymbol{z})= \frac{\exp{z_{t}}}{\sum_{j=1}^{T} \exp{z_{j}}} \text{ for $t\in\mathcal{T}$ and $\boldsymbol{z} \in \mathbb{R}^{|\mathcal{T}|}$}.
\label{eq:softmax}
\end{equation}

We obtain the final prescription of the network by finding the treatment $t$ with the highest probability $\mathbb{P}[\tau(\boldsymbol{x}_i) = t] = \sigma_t(\boldsymbol{z})$. This approach is analogous to a classification network, where the predicted class is the one with the highest probability among the network's output nodes. The tractable objective for our PNN models is therefore:
\begin{equation}
    \min_{\tau(.)} \frac{1}{n} \sum_{i=1}^{n} \sum_{t\in\mathcal{T}} \mathbb{P}[\tau(\boldsymbol{x}_i) = t] \cdot \Gamma_{i,t}.
\label{eq:prob_loss_func}
\end{equation}

\subsection{Recovering Interpretability with Optimal Classification Trees}\label{subsec:oct-pnn}

For structured datasets, we are able to recover interpretability through the use of knowledge distillation, in which we fit Optimal Classification Trees \citep{optimal-classification-trees,MLbook} on the feature data and prescription outputs of the PNN.
We present an example of such a Mirrored OCT in Figure \ref{fig:reboa_tree} (with other examples available in Appendix \ref{appendix-oct-pnn}). This example comes from the REBOA (resuscitative endovascular balloon occlusion of the aorta) application in Section \ref{subsec:unstructured}. In this real-world problem, we aim to minimize patient mortality by either prescribing (treatment 1) or not prescribing (treatment 0) the REBOA treatment. The tree in the figure is fit on the same observational data used to train its corresponding PNN, while the PNN prescriptions are used as target classes. 
We observe that this tree is very interpretable, and the features chosen for the splits come from our structured, observational data. If a patient (sample) is assigned to a leaf where the prediction is 0, then they are not prescribed the treatment, and if they are assigned to a leaf where the prediction is 1, the REBOA treatment is recommended.

\begin{figure}[!ht]
  \caption{Example of REBOA Mirrored OCT.}\label{fig:reboa_tree}
  \begin{center} 
  \includegraphics[width=0.8\textwidth]{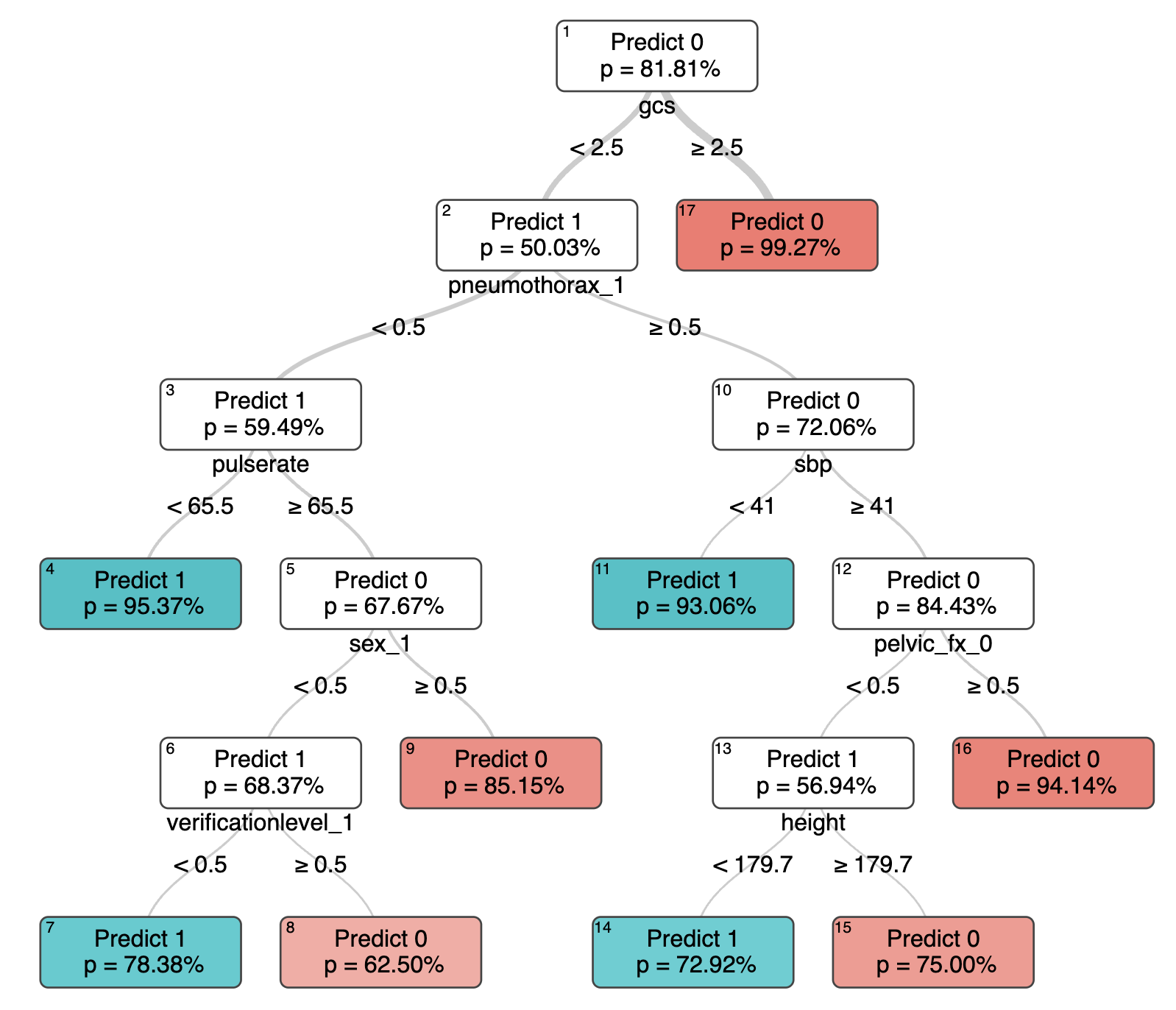}
  \end{center}
\end{figure}

\section{Experiments with real-world datasets}\label{sec:realworld-datasets}
In this section, we apply PNNs on real-world datasets that are both multimodal and unimodal. We first review methodology for data splits, network architecture, and performance evaluation, which are relevant for all of our experiments. We then report results and relevant discussions for each of our two multimodal datasets and four unimodal datasets.

\subsection{Data splits}
We split each dataset evenly, using 50\% for the training and 50\% for the test set. This choice is attributed to the fact that in order to evaluate the performance of the prescriptive methods on a test set, knowing the outcomes of the samples under the different treatments is required. Thus, counterfactual estimation is necessary for the test set too. To ensure fairness, and a high quality counterfactual estimation for the test set, enough test set points are necessary, and thus the typical ratios of 80-20 or 70-30 are not applicable in this case.

\subsection{Network Architecture}
We uniquely tune the PNN architecture for each dataset. We specify and finetune the following hyperparameters of the PNN:

\begin{itemize}
    \item \textbf{Number of layers of the network.} We experiment with both shallow and deep networks. Though conclusions differ based on the dataset, in general we observe that deeper networks do not necessarily improve results.
    \item \textbf{Number of nodes at each layer.} The size of the dataset closely affects this hyperparameter. Typically more nodes per layer are used with larger datasets that also include more features. 
    \item \textbf{Batch size.} This parameter determines the number of samples used in each forward pass of the network and for backpropagation, where the network parameters are updated after each batch passes through the network. It also affects the training speed, since too many batches can slow down the training process. Again, there is a correlation between batch size and the size of our data; larger batch sizes are employed for larger datasets.
    \item \textbf{Learning rate.} The learning rate is an important parameter of the training  process, since it defines how steep the descent is at each step of the gradient descent algorithm during training. After experimentation for each dataset, we find an appropriate learning rate that is not too big so that the algorithm becomes stuck in local optima but also not too small so that convergence is too slow.
    \item \textbf{Weight decay.} This hyperparameter scales an $L_2$-regularization term of the network weights that is added to the objective function to prevent them from taking too large values. Since our data is normalized, we observe that lowering the weight decay coefficient and relaxing the weights are actually beneficial and do not result in overfitting.
    \item \textbf{Number of epochs.} The number of epochs is a particularly hard parameter to tune, since we want to prevent overfitting but also allow the training to continue until sufficient convergence. For this reason, we employ early stopping, a technique that is adaptive to each specific training process and terminates training when a fluctuation in the validation loss is observed. Such a fluctuation indicates that the network is no longer improving in out-of-sample data generalization.
\end{itemize}
On top of these architectural choices, we choose the Adam optimizer. We tune the aforementioned hyperparameters using a validation set we extract from the training set that is not used in model training. We typically keep 15\% of the training data for the validation set. The Mirrored OCTs are then trained on the prescriptions from each of the PNN models.

\subsection{Experiments} \label{subsec:experiments}

To evaluate the models' performance on each dataset, we perform multiple train-validation-test splits and report the average performance of each split's models. This ensures that the results are not tailored to a specific data split, and also enables the investigation of stability of the prescriptive methods. Also, given the randomness often associated with training machine learning models, we train multiple models per split and also average their performance. In total, we perform 5 randomized data splits per dataset, and we train 5 models per split, so we train 25 models in total, for each model type.

\subsection{Performance metrics} \label{subsec:best_model}

To assess model performance, we use a relative outcome improvement metric measured on the unseen test set. This metric compares the estimated outcome of the treatment prescribed by the model with that of the real-life treatment. For most datasets, where doubly robust estimation is applied, test-set reward matrix entries lack natural meaning. Thus, instead of comparing with the actual outcome and to ensure fairness, hypothetical outcomes for both treatments (model-prescribed and real-life) are drawn from the reward matrix and then compared. The average relative outcome improvement is then computed as:

\begin{equation}
    \bar{I} = \frac{ \sum_{i=1}^{n} |\Gamma_{i,\hat{t}_{i}} - \Gamma_{i,t_{i}}|} {\sum_{i=1}^{n}\Gamma_{i,t_{i}}},
\end{equation}
where $\Gamma_{i,\hat{t}_{i}}$ is the estimated outcome for the $i^{\text{th}}$ observation if the prescribed treatment $\hat{t}_{i}\in\mathcal{T}$ is applied and $\Gamma_{i,t_{i}}$ is the estimated outcome for the $i^{\text{th}}$ observation under treatment $t_{i}$.

For the case of unstructured datasets, the evaluation on the test set can be performed using counterfactuals that have been calculated either by using a single modality, or multimodal data. To address that, the outcome improvement is evaluated using the test set counterfactuals with both types of models (in this case, tabular, and multimodal, from tabular and notes).

\subsection{Unstructured datasets} \label{subsec:unstructured}
We demonstrate the efficacy of our PNN models on two real-world, multimodal datasets: transcatheter aortic valve replcement (TAVR) and liver trauma injuries. Both datasets include tabular (structured) and clinical notes (unstructured) data, and we train two models: unimodal models -- fit on just the tabular data -- and multimodal models -- fit on the combined tabular and notes data. We report the results of both datasets and discuss the improved performance of the multimodal pipeline in this section. To ensure that there is no bias stemming from which modality is used to perform the reward estimation in the test set, we report the estimated outcome improvement under rewards estimated using both the single modality (tabular data) and the multiple modalities (tabular and notes). 

\textbf{Transcatheter aortic valve replacement (TAVR).} Transcatheter aortic valve replacement (TAVR) is a treatment option for patients with severe aortic stenosis across all levels of surgical risk. In the United States, two transcatheter heart valves (THV) are used, the balloon-expandable Edwards Sapien 3 and the self-expanding Medtronic Evolut Pro Plus. Selection of valve choice by medical professionals is generally based on several factors including operator preference, patient characteristics, and valvular/annular anatomy on a computerized tomography (CT) scan \citep{Mitsisi2022, leone2023prosthesis}. Despite improvement in TAVR devices, implant techniques, and operator experience, permanent pacemaker implantation (PPI) continues to remain a frequent complication with an estimated prevalence of 7-18\% \citep{Webb2012, smith2011transcatheter}, with potential consequences on patients' mortality and cost of care. This dataset contains demographic (e.g. age, sex, bmi) and medical information (e.g. hypertension, Left Ventricular Ejection Fraction), as well as radiology reports of echocardiograms and CT scans from 2,127 patients, and the problem we seek to solve is prescribe the most appropriate type of valve to patients, so that their risk of PPI is minimized. We train two different sets of models, one where only the tabular features are considered, and one where notes are also incorporated, in the form of embeddings, extracted as described in Section \ref{subsubsec: unstructured}.

\textbf{Liver trauma injuries.} Acute liver injury is considered one of the two most common solid organ injuries in blunt trauma victims. However, inaccuracies exist in the grading of liver injuries by human read and interpretation of CT scans, which may lead to mistreatment \citep{georg2014accuracy}. Therefore, personalized treatment for the patient is important in trauma management. This dataset comes from electronic medical records of 722 liver injury patients and includes features such as patient demographics, history of illness, lab results, and allergies. We aim to prescribe either surgical or non-surgical intervention to minimize patient mortality (binary outcome). 

The results for both datasets are presented in Table \ref{tab:liver-results}. For both of the datasets, we observe that the multimodal models significantly outperform the unimodal ones, by even more than 10\% on average, demonstrating the benefit of increased information from the added language modality. Although the discrepancy between the outcome improvement can be quite different depending on the modality used to train the test set reward estimators, there is a clear improvement observed when multimodality is employed under both estimators. The results are also stable, across 5 different data splits and 5 different models per split, and indicate the prescriptive power multimodality can offer. 

\begin{table}[ht]\caption{Improvement(\%) in risk for the experiments with unstructured data (TAVR, liver trauma), where lower risk is better. We report the average improvement and standard deviation across the five splits.}\label{tab:liver-results}
\begin{center}
\begin{tabular}{cccccccc}
& & \multicolumn{2}{c}{\textbf{TAVR models}} & \multicolumn{2}{c}{\textbf{Liver trauma models}} \\ 
\textbf{Estimator} & \textbf{Method} & \textbf{Tabular} & \textbf{Multimodal} & \textbf{Tabular} & \textbf{Multimodal} \\ \midrule
\multirow{2}{*}{Tabular} & PNN & $3.41 \pm 5.06$ & $8.66 \pm 6.80$  & $16.34 \pm 10.09$ & $32.50 \pm 12.40$ \\ 
& Mirrored OCT & $4.84\pm 6.24$ & $10.10\pm 9.02$  & $24.86\pm 9.94$ & $34.96\pm 6.80$\\ \midrule
\multirow{2}{*}{\begin{tabular}{@{}c@{}}Tabular \\ \& Notes \end{tabular}} 
& PNN  & $20.50 \pm 3.86$ & $32.15 \pm 7.52$ & $25.32 \pm 5.10$ & $40.30 \pm 14.70$\\ 
& Mirrored OCT & $\mathbf{21.33\pm 10.43}$ & $\mathbf{34.50\pm 9.79}$ & $\mathbf{37.46\pm 5.87} $& $\mathbf{45.11\pm 4.18}$ \\
\end{tabular}
\end{center}
\end{table}

\begin{table}[ht]
    \caption{Training accuracy(\%) of the Mirrored OCTs for the unstructured datasets. We report the average accuracy and standard deviation across the five splits. }
    \begin{center}
    \begin{tabular}{ccc}
         \textbf{Dataset} & \textbf{Tabular model} & \textbf{Multimodal model}\\ \midrule
         TAVR & $79.03\pm 3.09 $& $75.73\pm 0.69$\\
         Liver trauma & $87.66\pm 3.50$ & $84.07\pm 3.11$ \\
    \end{tabular}
    \label{tab:accuracy_octs_unstr}
    \end{center}
\end{table}

Mirrored OCTs, trained on the PNNs' prescriptions, result in an even larger outcome improvement compared to the PNNs, in both datasets. This is a dataset-specific observation, that demonstrates, however, that Mirrored OCTs do not generally result in performance decrease. The training accuracy of the OCTs on the prescriptions is presented in Table \ref{tab:accuracy_octs_unstr}, and we observe that it is higher in the tabular compared to the tabular and notes model. This is to be expected, since the tabular model is trained using fewer features and thus is easier to be mirrored by a classification model.

\subsection{Structured datasets} \label{res:structured}
We now apply our PNN models to four real-world, unimodal tabular datasets: diabetes management, groceries pricing, splenic injuries treatment, and REBOA in blunt trauma patients. Because these are purely tabular datasets, we are able to recover interpretability by fitting Mirrored OCT models. 

Table \ref{table:structured-datasets} outlines the treatment scenarios covered by these four datasets, the counterfactual estimation method employed, as well as the training accuracy of the Mirrored OCTs, which we note to be high for all datasets. We present results for all four structured datasets in Table \ref{table:structured-datasets}, where we directly compare PNNs and their Mirrored OCTs with the performance of other well-known, state-of-the-art prescriptive methods, including Optimal Policy Trees, Regress \& Compare, and Causal Forests. 

For Regress \& Compare, we typically train an XGBoost Regressor or Classifier (depending on the nature of the outcome). For this purpose, we append the actual treatment as a separate column in the observational data and we train the predictive model to predict the real-life outcome under the treatment. To select the best treatment for a new sample, we append each of the available treatments separately and we obtain the final outcome in each case. The treatment that results in the best outcome is selected.

\begin{table}[ht]
\caption{Treatment scenarios, data splits, and counterfactual estimation methods used for each structured dataset.}\label{table:structured-datasets}
    \begin{center}
    \begin{tabular}{@{}ccccc@{}}
         \textbf{Dataset} & 
         \begin{tabular}{@{}c@{}}\textbf{Treatment} \\ \textbf{scenario}\end{tabular}  & 
         \begin{tabular}{@{}c@{}}\textbf{Counterfactual} \\ \textbf{estimator}\end{tabular} & \begin{tabular}{@{}c@{}}\textbf{Training} \\ \textbf{Accuracy(\%) of} \\ \textbf{Mirrored OCT} \end{tabular}\\  
    \midrule
        \begin{tabular}{@{}c@{}}Groceries \\ pricing\end{tabular} & \begin{tabular}{@{}c@{}}Single \\ continuous\end{tabular}
        & \begin{tabular}{@{}c@{}}XGBoost \\ Classifier \\ (direct method)\end{tabular} & $98.03\pm 0.81$ \\
        \hline
        \begin{tabular}{@{}c@{}}Splenic injuries \\ treatment\end{tabular} & \begin{tabular}{@{}c@{}}Multiple \\ discrete\end{tabular}
        & \begin{tabular}{@{}c@{}}Random Forest \\ Classifier \\ (doubly-robust)\end{tabular} & $91.99\pm 1.65$\\
        \hline
        \begin{tabular}{@{}c@{}}REBOA in blunt \\ trauma patients\end{tabular}  & Binary & 
        \begin{tabular}{@{}c@{}}XGBoost \\ Classifier \\ (doubly-robust)\end{tabular} & $96.93\pm 0.69$\\
        \hline 
        \begin{tabular}{@{}c@{}}Diabetes \\ management\end{tabular}& \begin{tabular}{@{}c@{}}Multiple \\ continuous\end{tabular} & 
        \begin{tabular}{@{}c@{}}Random Forest \\ Regressor \\ (doubly-robust)\end{tabular} & $91.54\pm 3.07$ \\
    \end{tabular}
    \end{center}
\end{table}

\textbf{Diabetes management.} This dataset is based on electronic medical records of 58,200 patients with type 2 diabetes from 1999 to 2014 from the Boston Medical Center. It contains information regarding patient demographics, a timeseries of insulin levels, as well as current drug prescriptions. Patient treatments include combinations of insulin, metformin, and oral blood glucose regulation agents, and patient outputs are the resulting hemoglobin A1C measurements (continuous outcome), for which lower values are more optimal. 

\textbf{Groceries pricing.} For this study, we select the
publicly-available retail dataset ``The Complete Journey'' \citep{groceries-dataset, groceries-article}, which contains household-level transactions of many products over two years of 2,500 frequent-shopper households. We focus on one specific product, strawberries. The task here is to, given household demographics, prescribe optimal prices to strawberries with a binary outcome indicating if the household purchases strawberries or not after being assigned the strawberry price. The objective is to maximize revenue, where revenue is defined as the price if strawberries are purchased and zero otherwise (e.g., binary). After filtering the data to only the relevant households that had purchased strawberries at least once, the final dataset consists of 97,295 transactions. We impute strawberry prices for cases where strawberry-purchasing households did not purchase strawberries on that specific trip by using the mode of the strawberry prices on the most recent day prior to the trip on which no strawberries were purchased. We consider prices from \$2 to \$5, inclusive, in increments of \$0.50. Since there does not seem to be a strong correlation between strawberry price and the covariate features, rewards are estimated using the direct method. 

\textbf{Splenic injuries treatment.} The spleen is an immunologic intra-abdominal organ on the left side of the body, which may be removed in the case of injury. In the 1970's to 1980's, the medical community saw a shift towards preservation of the spleen rather than removal, thus making it important to correctly determine if spleen removal was indeed necessary. This specific dataset includes data on spleen surgical operations, in addition to demographic and medical data consisting of numerical, binary, and categorical types. After preprocessing, we have 35,954 rows of patient data in this dataset. We aim to optimally prescribe splenectomy, angioembolization, or observation in blunt splenic injuries to minimize patient mortality (binary outcome).

\textbf{REBOA in blunt trauma patients.} The use of resuscitative endovascular balloon occlusion of the aorta (REBOA) for control of noncompressible torso hemorrhage continues to be highly debated. Being able to appropriately determine if such a treatment should be used is critical in order to decrease the misuse of the treatment in hemodynamically unstable blunt trauma patients. This dataset includes 9,998 patients, with features that are both demographic and medical in nature, including numerical, binary, and categorical values. The goal is to prescribe the REBOA treatment or not to minimize patient mortality (binary outcome). Some feature columns contain unknown values; we therefore use Optimal Imputation \citep{Bertsimas2017FromPM} with K-Nearest Neighbors to fill the missing values. A few features are integral, and we round imputed values to the nearest integer to maintain integrality.

As demonstrated in Table \ref{table:structured-datasets}\footnote{For the groceries dataset, improvement is computed as mean revenue improvement rather than outcome improvement, where mean revenue is $\Bar{p}_r = \frac{1}{n} \sum_{i=1}^{n} \Gamma_{i,\hat{t}_{i}} \cdot \hat{t}_{i}$, and actual revenue is  $\Bar{p}_r = \frac{1}{n} \sum_{i=1}^{n} y_{i,t_{i}} \cdot t_{i}$, $\hat{t}_i$ is the prescribed treatment for sample $i$ by the model, and $t_{i}$ is the real-life prescribed treatment.}, we generally see an improvement in estimated outcome across all methods for all four structured datasets. Notably, we observe that PNNs and the Mirrored OCTs perform very strongly in all experiments. In particular, the PNNs outperform the other models in the diabetes, groceries and spleen datasets, while performing comparably to Optimal Policy Trees in the REBOA dataset. In especially the diabetes dataset, PNNs remarkably outperform other models by roughly twice as much; since the diabetes dataset falls under the multiple continuous treatment case, the exceptional PNN performance reflects its ability to handle more complicated data. We also note that the Mirrored OCTs also generally perform very similarly to PNNs, with the added benefit of interpretability. 

\begin{table}[ht]\caption{Improvement(\%) for structured datasets. We report the average improvement and standard deviation across the five splits.}\label{table:structured-results}
\begin{center}
\begin{tabular}{ccccc}
\textbf{Method} & \textbf{Diabetes} & \textbf{Groceries} & \textbf{Spleen} & \textbf{REBOA} \\ \midrule
Regress \& Compare & $2.00\pm 0.25$ & $94.17 \pm 6.25$ & $8.03 \pm 2.24$ & $-17.80 \pm 27.91$ \\ 
Causal Forest & $ 2.50 \pm 0.69$& $98.68 \pm 5.98$ & $-4.84 \pm 13.14$& $-4.82 \pm 3.18$\\ 
Optimal Policy Tree & $1.72\pm 1.53$ & $106.58 \pm 5.33$ & $16.41 \pm 3.15$ & $\mathbf{10.79 \pm 4.34}$\\ 
PNN & $\mathbf{4.01 \pm 0.90} $& $\mathbf{110.88 \pm 5.91}$ & $\mathbf{20.47 \pm 1.68}$& $10.22 \pm 4.55$\\ 
Mirrored OCT & $3.93\pm 1.21$ & $110.22\pm 6.94$ & $15.72\pm 6.01$ &$10.21\pm 5.03$ \\ 
\end{tabular}
\end{center}
\end{table}

\section{Discussion}\label{sec:real-world-discussion}
In general, reporting improvement based on estimated rewards is a good approximation for evaluating the performance of prescriptive methods. However, such metrics do not provide any insights into how realistic the prescriptions are or how different they are from policies observed in the data. Another critical aspect is the models' stability. The prescriptive models should be robust in their prescriptions across dataset splits and with respect to inherent randomness during training. Furthermore, interpretability is crucial for real-world deployment, as users of models must understand where decisions are coming from in order to implement them. We therefore discuss these three topics -- realistic and stable prescriptions, as well as interpretability -- in the following sections. 

\subsection{Realistic Nature of Prescriptions}
Providing realistic prescriptions is crucial, particularly when employing prescriptive tools in practice. To evaluate the realism of the provided prescriptions, we quantify the deviation between the prescribed and real-life treatments of individual samples. This evaluation is carried out per model, by calculating the mean absolute difference between each sample's prescribed and real-life treatment throughout the dataset (training, validation, and test sets) and then averaging it across all samples. For discrete cases (REBOA and spleen datasets), the treatments are ordered in terms of severity, so that the distance makes sense as a metric. For this purpose, the $N_m=25$ trained models from each dataset are considered. The mean absolute difference for the $k$-th individual model is given by:

\begin{equation}
    D_k = \frac{1}{n}  \sum_{i=1}^n \| \hat{t}_i - t_i \|_1,
\end{equation}
where $n$ is the size of the dataset, $\hat{t}_i$ is the prescribed treatment for sample $i$, and $t_i$ is the treatment sample $i$ got in real life. The mean absolute difference across the $N_m=25$ models is then computed as:

\begin{equation}
    \bar{D} = \frac{1}{N_m} \sum_{k=1}^{N_m} D_k.
\end{equation}

The results are presented in Tables \ref{structured-mad} and \ref{unstructured-mad}. Clearly, we prefer both high performance $\bar{I}$ and low mean absolute difference $\bar{D}$, as this ensures that improvements in outcomes are not achieved through disproportionate shifts in treatment assignments. We naturally expect that PNN-prescribed treatments are more different than those in real-life, as compared to the other prescriptive methods, since as presented in Sections \ref{subsec:unstructured} and \ref{res:structured}, PNNs outperform the other methods in most datasets. Contrary to our expectation, however, we favorably observe that PNNs, as well as the Mirrored OCTs, result in mean absolute difference between the prescribed and the actual treatments that is comparable to the rest of the methods.

\begin{table}[ht]\caption{Mean Absolute Difference between prescribed and actual treatments for structured datasets.}\label{structured-mad}
\begin{center}
\begin{tabular}{ccccc}
\textbf{Method} & \textbf{Diabetes} & \textbf{Groceries} & \textbf{Spleen} & \textbf{REBOA} \\ \midrule
Regress \& Compare & $\mathbf{0.4556}$ & $1.033$ & $\mathbf{0.3355}$ &  $0.2815$ \\ 
Causal Forest & $0.5923$& $\mathbf{1.0042}$ & $ 0.7761$ & $0.2492$\\ 
Optimal Policy Tree & $0.7490$&$1.0221$ & $0.3755$ &  $\mathbf{0.1199}$\\ 
PNN & $0.4903$ & $1.4386$ & $0.5185$ &  $0.1568$\\ 
Mirrored OCT & $0.4936$ & $1.4391$ & $0.5069$ &  $0.1435$\\ 
\end{tabular}
\end{center}
\end{table}

\begin{table}[ht]\caption{Mean Absolute Difference between prescribed and actual treatments for unstructured datasets.}\label{unstructured-mad}
\begin{center}
\begin{tabular}{ccccccc}
& \multicolumn{2}{c}{\textbf{TAVR models}} & \multicolumn{2}{c}{\textbf{Liver trauma models}} \\ 
\textbf{Method} & \textbf{Tabular} & \textbf{Multimodal} & \textbf{Tabular} & \textbf{Multimodal} \\ \midrule
PNN & $0.4561$ & $\mathbf{0.6144}$  & $0.3590$ & $0.3837$ \\ 
Mirrored OCT & $\mathbf{0.4280}$ & $0.6539$ & $\mathbf{0.3351}$ & $\mathbf{0.3089}$\\ 
\end{tabular}
\end{center}
\end{table}

For unstructured data, Section \ref{subsec:unstructured} highlights the significant advantage of the multimodal approach over tabular-only methods. As expected, this performance edge indicates that the proportion of prescription changes is higher in the multimodal case. The results indicate that, in the TAVR case for example, around 62\% of the PNN prescriptions change from one valve to the other in the multimodal case, and 45\% in the tabular case, which is a considerable shift. For the liver trauma dataset, the difference between the prescribed and the actual treatments is smaller. 

A critical advantage of neural networks is that the user has some control over how much the prescriptions change. For example, depending on the application, a threshold can be selected that limits the number of treatment assignment modifications, and only models that satisfy this constraint on the validation set are considered. Alternatively, one can incorporate a penalty term in the objective function, to penalize an excessive number of treatment switches. This is application-specific, but highlights the flexibility that neural networks offer compared to other prescriptive methods.

The realism of prescriptions is also evaluated by examining the average number of distinct prescriptions per type of model, which shows how much the model is capable of utilizing the full treatment space. This is evaluated as:

\begin{equation}
    \bar{N} = \frac{1}{|\mathcal{T}|} \cdot \frac{1}{N_m} \sum_{i=1}^{N_m}| t: \exists j: \hat{t}_j = t, \ j=1,\dots, n|,
\end{equation}
where the average number of prescriptions is normalized by the size of treatment space, to calculate a percentage and thus make the metric comparable across the different datasets. The results are presented in Tables \ref{structured-perc} and \ref{unstructured-perc}.

\begin{table}[ht]\setlength\extrarowheight{1.35pt}\caption{Percentage of different prescriptions selected by the models for structured datasets.}\label{structured-perc}
\begin{center}
\begin{tabular}{ccccc}
\textbf{Method}
& \textbf{Diabetes} & \textbf{Groceries} & \textbf{Spleen} & \textbf{REBOA} \\ \midrule
Regress \& Compare & $36.92$ &  $23.33$ & $66.67$ & $70.0$ \\ 
Causal Forest & $99.69$& $95.33$ & $ 88.00$ & $100.0$\\ 
Optimal Policy Tree & $59.38$ &  $48.67$ & $85.33$ & $100.0$\\ 
PNN & $16.62$ & $49.33$ & $81.33$& $100.0$\\ 
\end{tabular}
\end{center}
\end{table}

\begin{table}[ht]\caption{Percentage of different prescriptions selected by the models for unstructured datasets.}\label{unstructured-perc}
\begin{center}
\begin{tabular}{ccccccc}
& \multicolumn{2}{c}{\textbf{TAVR models}} & \multicolumn{2}{c}{\textbf{Liver trauma models}} \\ 
\textbf{Method} & \textbf{Tabular} & \textbf{Multimodal} & \textbf{Tabular} & \textbf{Multimodal} \\ \midrule
PNN & $100.0$ & $100.0$  & $100.0$ & $100.0$ \\ 
Mirrored OCT & $100.0$ & $98.0$ & $86.0$ & $76.0$\\
\end{tabular}
\end{center}
\end{table}

In most of the structured datasets, we observe that PNNs and Mirrored OCTs prescribe a high percentage of the available treatments, with the exception of the diabetes dataset, where the treatment options are multiple. The Regress \& Compare approach clearly underutilizes the treatment space, since it prescribes a small percentage of treatments in most cases; this reveals potential treatment assignment bias that may not be mitigated through this approach. Causal Forests seem to make prescriptions that mostly cover the full treatment regime; however their performance is worse than Optimal Policy Trees and PNNs, as discussed in Section \ref{res:structured}. In the unstructured case, both tabular and multimodal PNN models employ all of the available treatments. We observe that some of the Mirrored OCTs only prescribed one treatment in the liver dataset, but most of them prescribed both. Naturally, as the multimodal datasets are more complex, the resulting networks are harder to be explained by trees.

Most importantly, PNNs are flexible in this feature too; by using the dropout mechanism \citep{srivastava2014dropout} in the last layers of the network, which is often used to prevent overfitting in neural networks, all of the output nodes are forced to be activated during training. As a result, more areas of the network are used, which empirically shows an increase in selected treatments by the model. The flexibility of PNNs is also underlined by the fact that they provide, for each observation, a probability of each treatment, similarly to a classification problem, where neural networks provide a probability of each class. The results presented for PNNs consider the prescription with the highest probability for each sample. However, one can employ a treatment-specific probability threshold to select the final treatment, like in classification problem; for example, this may be done according to some predefined, meaningful, treatment allocation percentage. This provides the user with some control over the resulting treatment distribution.

Overall, PNNs achieve a balance between performance and realism in prescriptions, and they also reasonably cover the treatment space. These are important factors that make them reliable for practitioners and their leverage in different real-life applications.
\subsection{Stability}

Given the randomness that is present when training neural networks, their stability compared to other machine learning models is often criticized \citep{colbrook2022difficulty}. The goal of this section is to compare the stability between the different prescriptive approaches by measuring the standard deviation of each observation's treatment distribution, which results from the $N_m=25$ different models that have been trained for each dataset. Ideally, the prescriptions should be consistent across the different model runs and data splits; otherwise the method is very sensitive to the training data distribution, which reduces the credibility of the prescriptions. 

For each observation, the standard deviation of its prescriptions across the different $N_m$ models is calculated, and we present averages across the different observations in Tables \ref{structured-stab}\footnote{For multiple continuous treatments (diabetes dataset), to get the standard deviation for each sample, we first calculate the standard deviation for each drug separately and then we average across the three drugs. For Regress \& Compare, since we use XGBoost models, there is no randomness in each split, so the 5 models produce the same prescriptions. This explains why in most of the datasets, Regress \& Compare has the lowest standard deviation.} and \ref{unstructured-stab}.

\begin{table}[ht]\caption{Standard deviation of each sample's prescriptions distribution across $N_m=25$ models for structured datasets.}\label{structured-stab}
\begin{center}
\begin{tabular}{ccccc}
\textbf{Method}
& \textbf{Diabetes} & \textbf{Groceries} & \textbf{Spleen} & \textbf{REBOA} \\ \midrule
Regress \& Compare & $\mathbf{0.0752}$ & $\mathbf{0.0285}$ & $\mathbf{0.0940}$ & $0.4230$ \\ 
Causal Forest & $0.5139$ & $ 0.1667$ & $0.4535$ & $0.3838$\\ 
Optimal Policy Tree &$0.5911$& $0.7227$ & $0.1877$ & $\mathbf{0.0225}$\\ 
PNN & $0.2643$& $0.9332$ & $0.3178$& $0.0679$\\ 
\end{tabular}
\end{center}
\end{table}

\begin{table}[ht]\caption{Standard deviation of each sample's prescriptions distribution across $N_m=25$ models for unstructured datasets.}\label{unstructured-stab}
\begin{center}
\begin{tabular}{ccccccc}
& \multicolumn{2}{c}{\textbf{TAVR models}} & \multicolumn{2}{c}{\textbf{Liver trauma models}} \\ 
\textbf{Method} & \textbf{Tabular} & \textbf{Multimodal} & \textbf{Tabular} & \textbf{Multimodal} \\ \midrule
PNN & $\mathbf{0.4062}$ & $\mathbf{0.4239}$  & $0.3336$ & $0.3202$ \\ 
Mirrored OCT & $0.4563$ & $0.4387$ & $\mathbf{0.2190}$ & $\mathbf{0.1032}$\\
\end{tabular}
\end{center}
\end{table}

We observe that the standard deviation of PNNs' prescriptions is comparable to the other models across all datasets, which indicates that although training neural networks is associated with inherently more randomness (random weight initialization, stochastic gradient descent), they result in relatively consistent prescriptions across different data splits and different runs, similarly to the more deterministic prescriptive methods. 

In particular, for the unstructured datasets in Table \ref{unstructured-stab}, the standard deviation of Mirrored OCTs is considerably smaller than the PNNs' in the liver trauma dataset. We attribute this to the fact that Mirrored OCTs were likely not able to fully capture the PNNs' complexity and are therefore simpler in their decision-making rules. For the TAVR dataset, the deviation is consistent across the different modalities, which indicates that adding a modality does not impact the model's stability.

\subsection{Interpretability}\label{subsec:interpretability}
We discuss now the interpretability of PNNs that is recovered via knowledge distillation of the Mirrored OCTs. In particular, we discuss the unstructured TAVR dataset, for which we can partially recover interpretability, and for the structured diabetes management dataset. Please refer to Appendix \ref{appendix:interpretability-continued} for similar analyses for our other datasets and Appendix \ref{appendix-oct-pnn} for model visualizations. 

\textbf{TAVR.} We discuss the multimodal Mirrored OCT from Figure \ref{fig:tavr-octpnn} for the TAVR dataset. While the embedding features from the clinical notes are not interpretable, we can still recover some interpretability through the tabular features selected by the Mirrored OCT. We see that the OCT selects one note feature, as well as the Valve-to-Annular Aortic Valve Area ratio (VDAoVA), age, and difference between the annular area of the patient's native aortic valve and the area of the prosthetic valve being implanted (Area Oversize). We observe that the OCT provides an insight into the prescriptions made by the PNN, even when it is trained on the multimodal data.

\textbf{Diabetes management.} We consider and compare the Mirrored OCT from Figure \ref{fig:diabetes-octpnn} and the Optimal Policy Tree from Figure \ref{fig:diabetes-opt} for diabetes management. Both models were trained with the same data, methods, and parameters as in Section \ref{res:structured}. For visualization reasons, Figure \ref{fig:diabetes-octpnn} displays a portion of the tree. From Figure \ref{fig:diabetes-octpnn}, we can see that the features selected by the tree include ``HbA1c\_mean'' (average pre-prescription blood hemoglobin A1C level), ``visitNo'' (number of visits by the patient), and ``pastHbA1c1'' (past blood hemoglobin A1C level). For example, if a patient has an average pre-prescription blood hemoglobin A1C level of less than 8.582, has at least 4 visits, and has past hemoglobin A1C levels of less than 9.15, then  the patient would be prescribed with treatment ``4,'' which corresponds to 0 units of insulin, 1 unit of metformin, and 0 units of oral blood glucose regulation agents. The Optimal Policy Tree (Figure \ref{fig:diabetes-opt}) is simpler, but equally interpretable. We observe that it selects different features; for example, the patient's average and past blood hemoglobin A1C levels are not taken into account. 

\section{Conclusions}
With its classification-like feedforward neural network architecture, our PNN framework flexibly handles multimodal data, by easily enabling the incorporation of multiple data sources. Furthermore, it is widely applicable for all treatment scenarios, and has the potential of making a significant impact in a variety of settings, as shown in our extensive real-world experiments. To the best of our knowledge, our method is the first prescriptive model for multimodal data, and it also outperforms or performs comparably to other well-known prescriptive models on unimodal tabular data in all treatment scenarios, without requiring large computational resources. 

Our approach is not only shown to perform strongly quantitatively, but also to provide realistic and stable prescriptions. The discrepancy between the prescribed and real-life treatment distributions is comparable to the other prescriptive methods. The small standard deviation of each sample's assignments from the models indicates that the networks are stable and robust to different data splits. Also, PNNs offer the advantage of flexibility since the user can adjust the loss function to provide partial control to the prescriptions, leveraging expert knowledge. 

Deep learning methods generally sacrifice interpretability. On unimodal tabular datasets, we are able to recover interpretability through a knowledge distillation approach leveraging interpretable OCT models, and on multimodal datasets, some interpretability may still be recovered. These Mirrored OCTs demonstrate similarly high performance in our real-world experiments, demonstrating that we can maintain high performance without sacrificing interpretability. This recovery of interpretability is critical for real-world deployment of deep learning models.

We conclude that our multimodal deep learning framework, PNNs, offers both flexibility and strong performance, effectively utilizing deep learning to process multimodal data. By integrating multiple data sources, the framework significantly enhances decision-making capabilities. This unified approach demonstrates its potential as a versatile prescriptive tool, well-suited for a wide range of applications.

\bibliography{main}
\bibliographystyle{tmlr}

\appendix
\section{Appendix}

\subsection{Extension of interpretability of unstructured and structured real-world datasets}\label{appendix:interpretability-continued}

\begin{sidewaysfigure}
  \begin{center}
    \caption{Mirrored OCT of maximum depth 7 for liver trauma.}\label{fig:liver-octpnn} \includegraphics[width=0.45\textwidth] {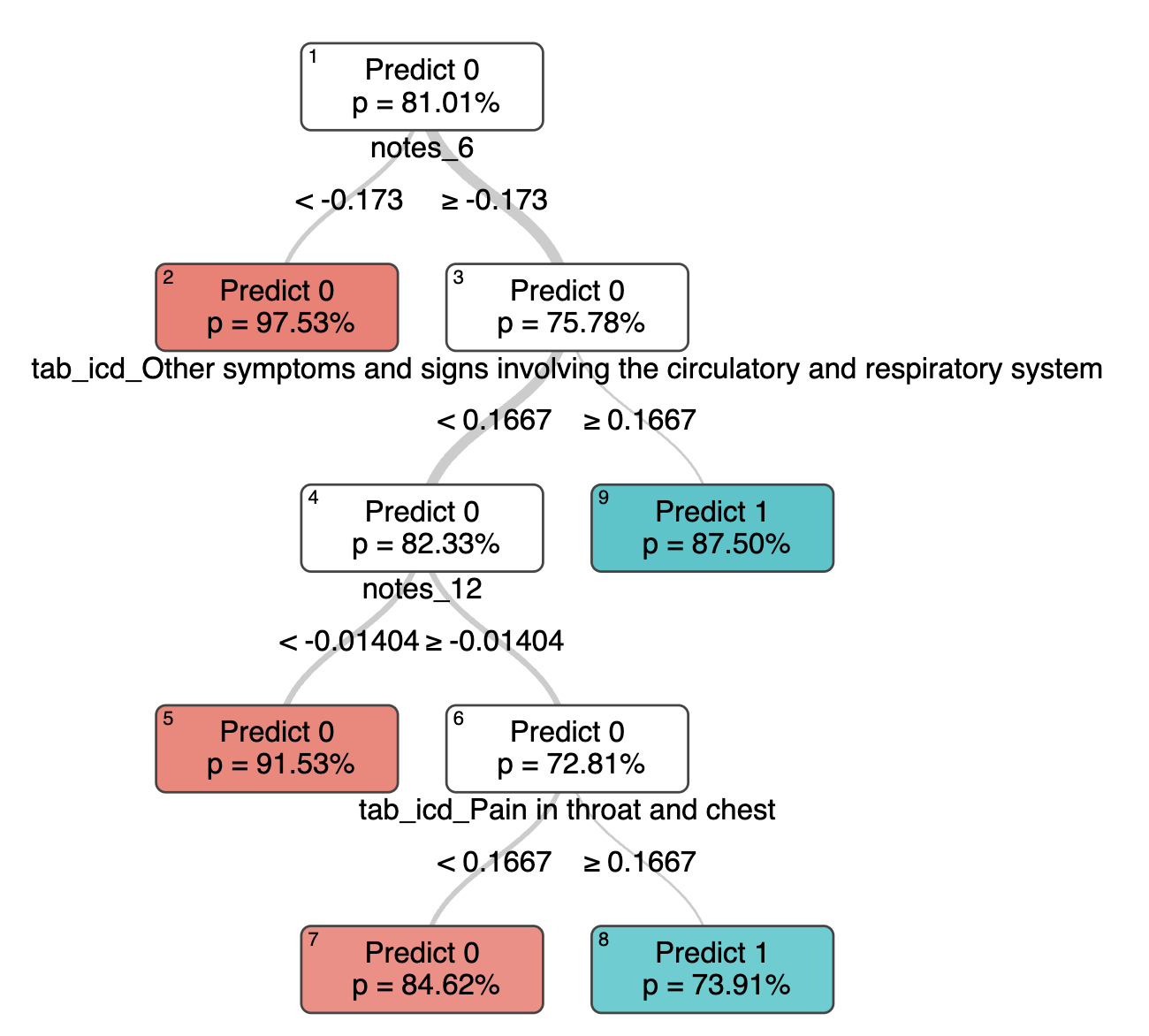}
    \end{center}
\end{sidewaysfigure}

\textbf{Liver trauma.} We discuss the multimodal Mirrored OCT from Figure \ref{fig:liver-octpnn} for the liver injury dataset. The embedding features from the clinical notes are not nearly as interpretable, but we can still interpret the tabular features selected by the Mirrored OCT. We see that the OCT selects two note features, as well as whether the patient had other symptoms with the circulatory or respiratory system and whether the patient had pain in their throat or chest.

\begin{sidewaysfigure}
  \begin{center}
    \caption{Mirrored OCT of maximum depth 7 for groceries.}\label{fig:groceries-octpnn} \includegraphics[width=0.95\textwidth,height=0.35\textwidth] {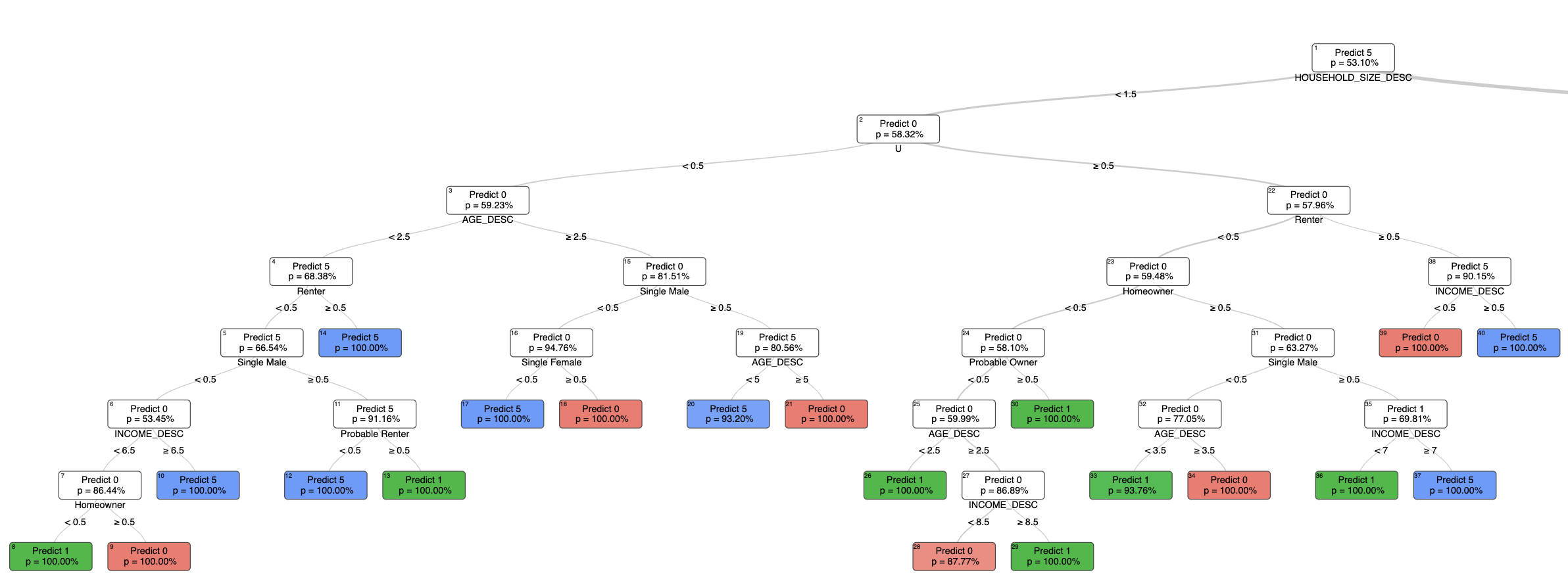}
    \end{center}
\end{sidewaysfigure}

\begin{sidewaysfigure}
  \begin{center}
    \caption{Optimal Policy Tree of maximum depth 7 for groceries.}\label{fig:groceries-opt}
  \includegraphics[width=0.95\textwidth,height=0.3\textwidth]{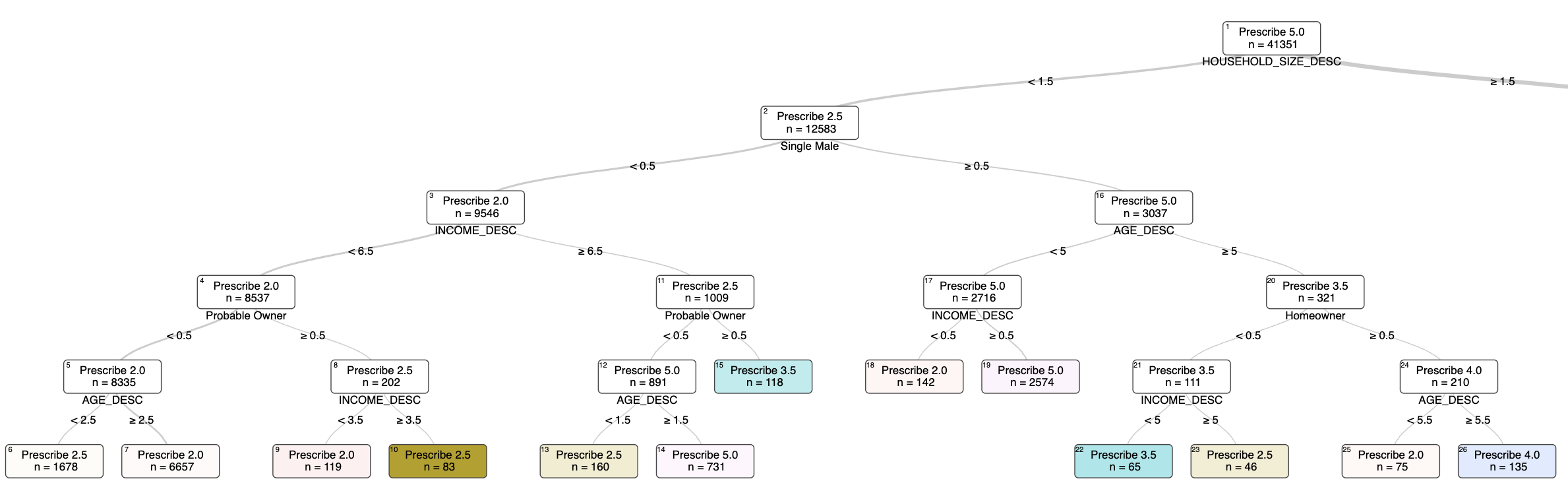}
  \end{center}
\end{sidewaysfigure}

\textbf{Groceries.} We compare the resulting Mirrored OCT from Figure \ref{fig:groceries-octpnn} and the Optimal Policy Tree (OPT) from Figure \ref{fig:groceries-opt}. 
The features selected by both the Mirrored OCT and the Optimal Policy Tree are similar, with the most prominent ones being the homeowner status, age, income range, household status, and marital status. An interesting difference between the two trees is the number of distinct prescriptions selected: the OCT only selects 3 of the 6 pricing options -- prediction classes 0, 1, and 5 (which correspond to prices USD \$2.00, \$2.50, and \$5.00) -- whereas the OPT prescribes more of the possible options. The OCT's strategy seems to therefore select low prices for lower-income households, compared to high prices for households with more financial stability. 
 
\begin{sidewaysfigure}
  \begin{center}
    \caption{Mirrored OCT of maximum depth 7 for spleen injury treatment.}\label{fig:spleen-octpnn} \includegraphics[width=0.95\textwidth,height=0.4\textwidth] {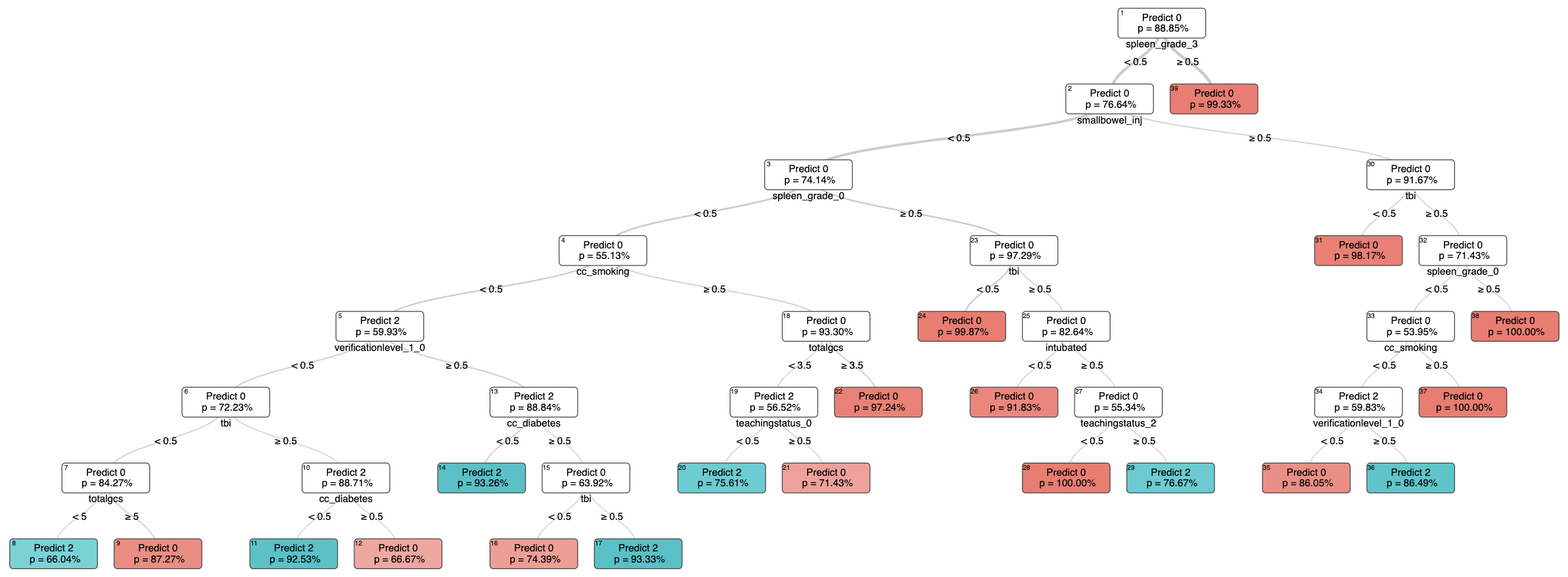}
    \end{center}
\end{sidewaysfigure}

\begin{sidewaysfigure}
  \begin{center}
    \caption{Optimal Policy Tree of maximum depth 7 for spleen injury treatment.}\label{fig:spleen-opt}
  \includegraphics[height=0.35\textwidth]{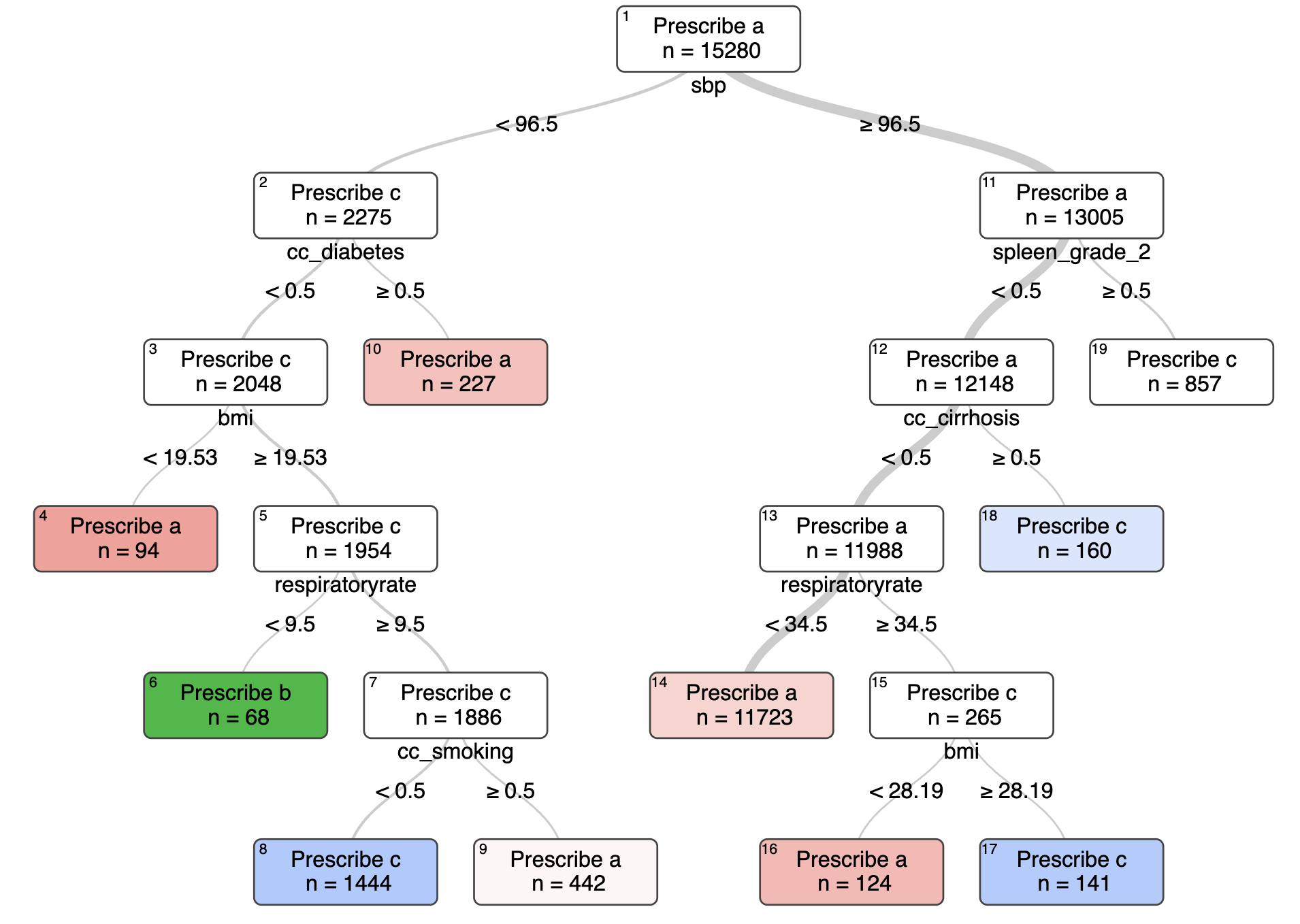}
  \end{center}
\end{sidewaysfigure}

\textbf{Splenic injuries treatment.} We compare the Mirrored OCT displayed in Figure \ref{fig:spleen-octpnn} with an Optimal Policy Tree (OPT) in Figure \ref{fig:spleen-opt}. 
There are three possible treatments: simple observation (treatment 0 of the OCT, ``a'' for the OPT), splenectomy (treatment 1 of the OCT, ``b'' for the OPT) and angioembolization (treatment 2 of the OCT, ``c'' for the OPT). The OCT prescribes the first and third options while the OPT prescribes all three. The two trees split on similar features, although at different levels of the tree; these features include ``totalgcs'' (Glasgow coma scale), the state (grade) of the spleen (injury severity on a scale of increasing acuteness from 1 to 5), diabetes, and smoking status. However, the Mirrored OCT splits on SBP (systolic blood pressure),  TBI (Traumatic brain injury) and intubation status, while the OPT splits on BMI, respiratory rate, and history of cirrhosis.

\begin{sidewaysfigure}
  \begin{center}
    \caption{Mirrored OCT of maximum depth 7 for REBOA in blunt trauma patients.}\label{fig:reboa-octpnn}
  \includegraphics[width=0.45\textwidth]{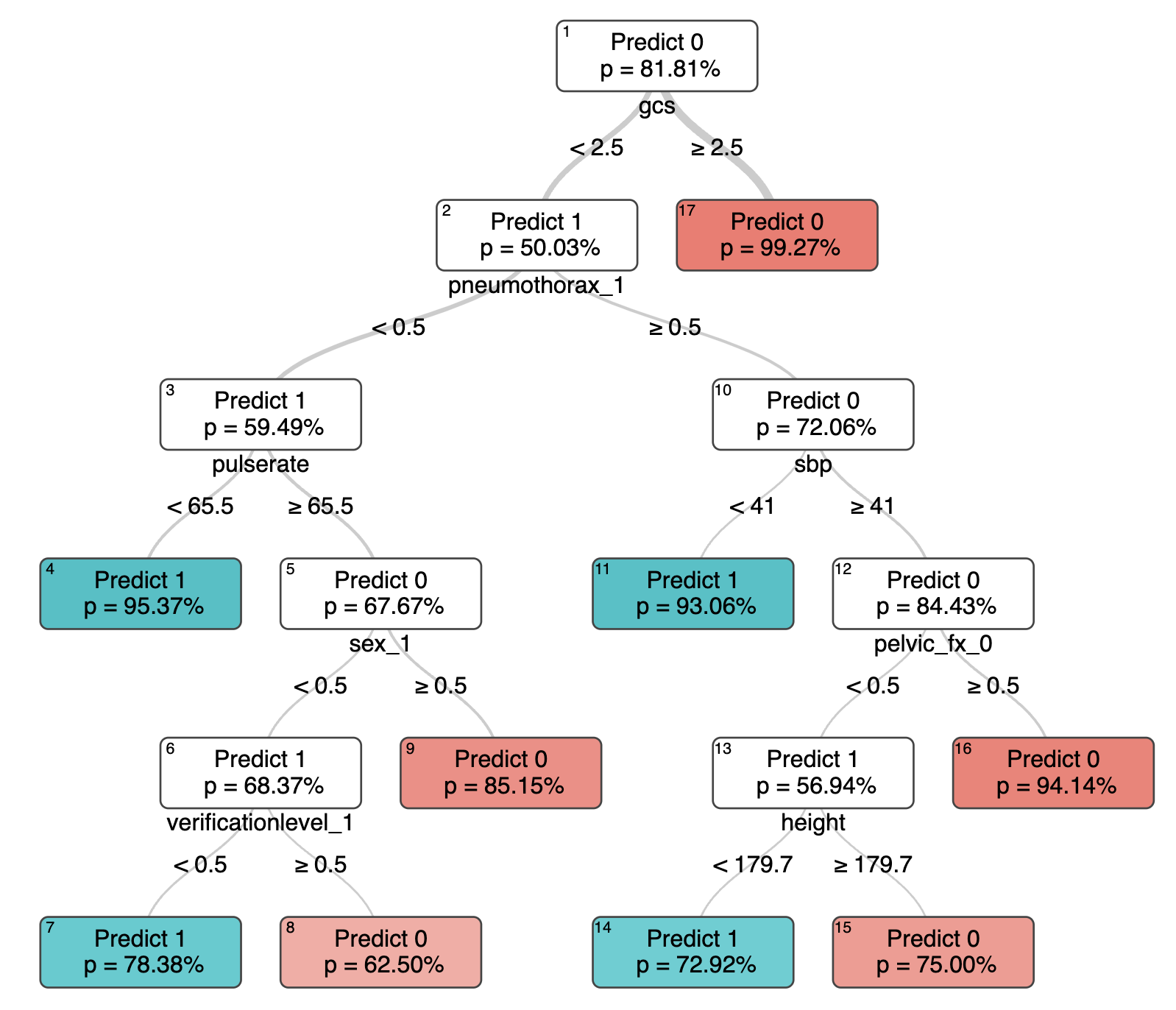}
  \end{center}
\end{sidewaysfigure}

\begin{sidewaysfigure}
  \begin{center}
    \caption{Optimal Policy Tree of maximum depth 7 for REBOA in blunt trauma patients.}\label{fig:reboa-opt}
  \includegraphics[width=0.35\textwidth]{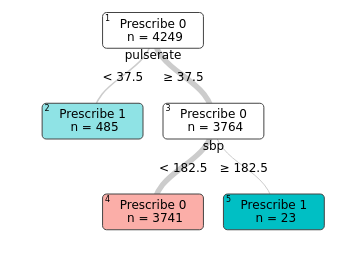}
  \end{center}
\end{sidewaysfigure}

\textbf{REBOA in blunt trauma patents.} We compare the Mirrored OCT from Figure \ref{fig:reboa-octpnn} and the Optimal Policy Tree (OPT) from Figure \ref{fig:reboa-opt}. 

From Figure \ref{fig:reboa-octpnn}, we clearly see that the important features include, among others, ``pneumothorax\_1'' (whether the patient has pneumothorax, a collection of air outside the lung but within the pleural cavity), ``gcs'' (Glasgow coma scale), SBP (systolic blood pressure), and pulse rate. We observe that the OPT (Figure \ref{fig:reboa-opt}) only splits on SBP and pulse rate, indicating that the resulting Mirrored OCT is considering more factors to make the final prescription, which is potentially closer to real life scenarios. 

\subsection{Mirrored OCTs on all real-world datasets}\label{appendix-oct-pnn}
Further examples of Mirrored OCTs of maximum depth 7 for the TAVR, liver trauma, diabetes, groceries, splenic injuries, and REBOA datasets can be found under \url{https://drive.google.com/drive/folders/12XNOQllyzVQEFFguHvkcAQ-7Psp9KriQ}.

\textcolor{white}{Force Appendix above figures}

\begin{sidewaysfigure}
  \begin{center}
    \caption{Mirrored OCT of maximum depth 7 for TAVR.}\label{fig:tavr-octpnn} \includegraphics[height=0.45\textwidth] {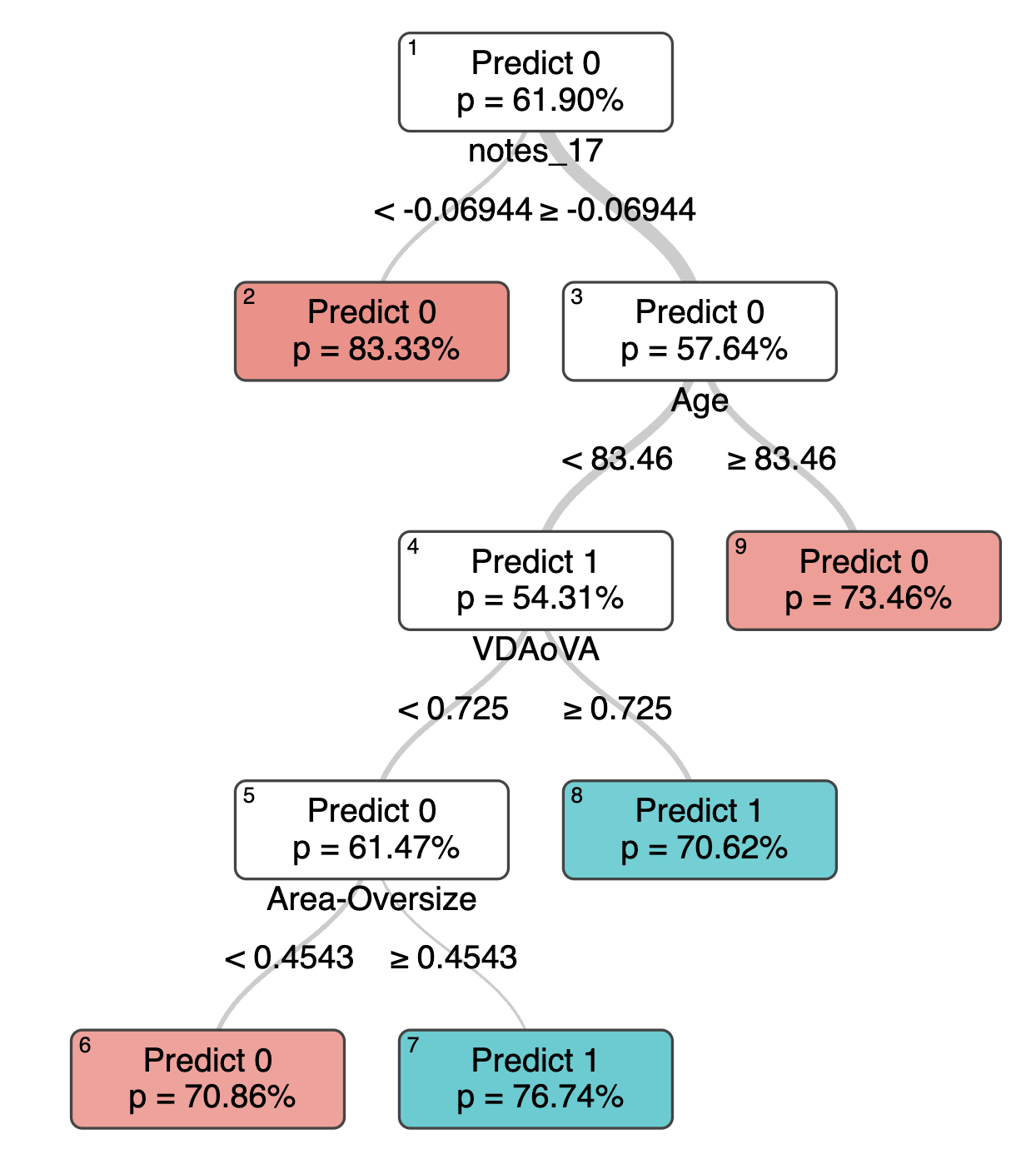}
    \end{center}
\end{sidewaysfigure}

\textcolor{white}{Force Appendix above figures}

\textcolor{white}{Force Appendix above figures}

\begin{sidewaysfigure}
  \begin{center}
    \caption{Mirrored OCT of maximum depth 7 for diabetes management.}\label{fig:diabetes-octpnn} \includegraphics[width=0.95\textwidth,height=0.5\textwidth] {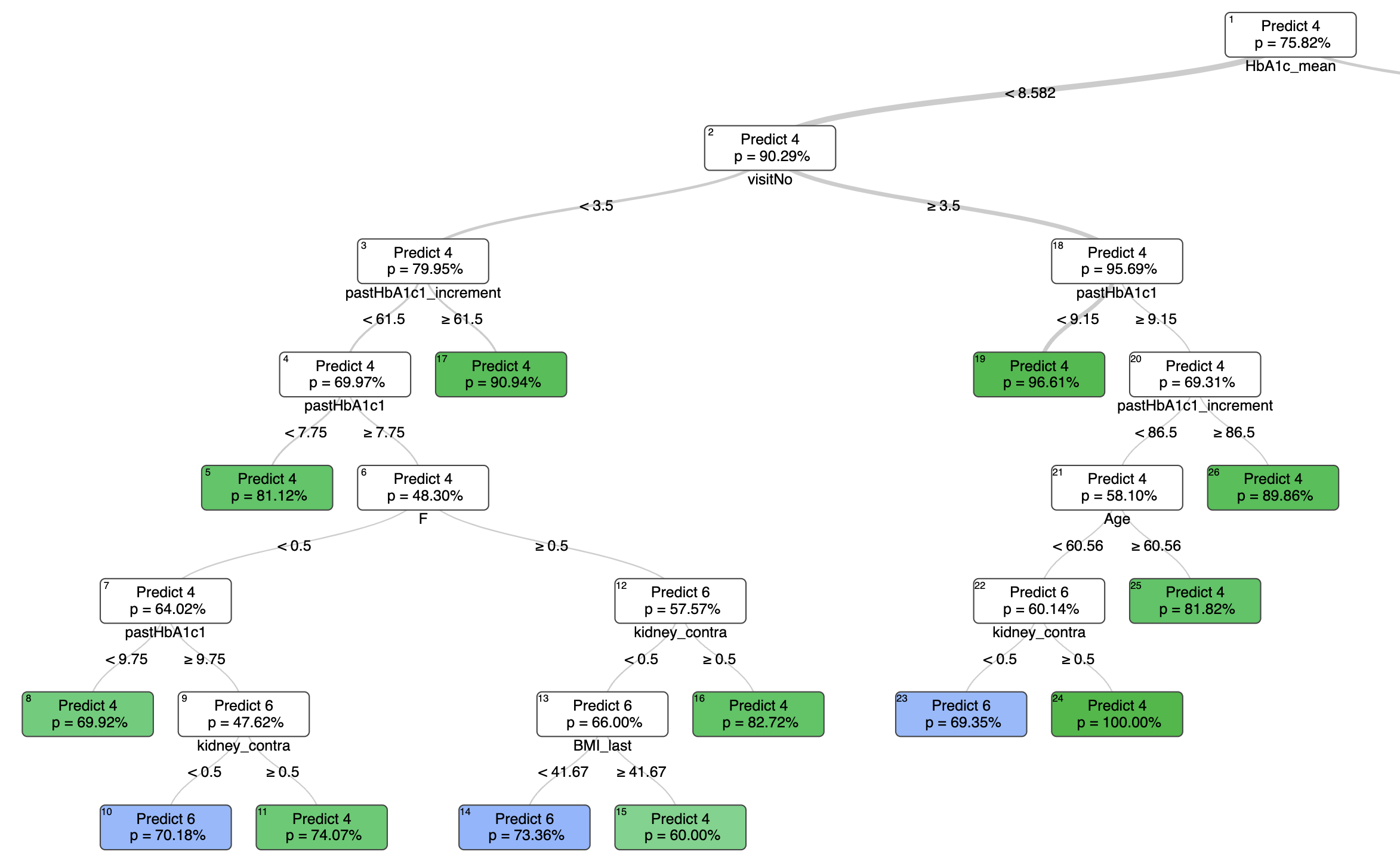}
    \end{center}
\end{sidewaysfigure}

\begin{sidewaysfigure}
  \begin{center}
    \caption{Optimal Policy Tree of maximum depth 7 for diabetes management.}\label{fig:diabetes-opt}
  \includegraphics[height=0.25\textwidth]{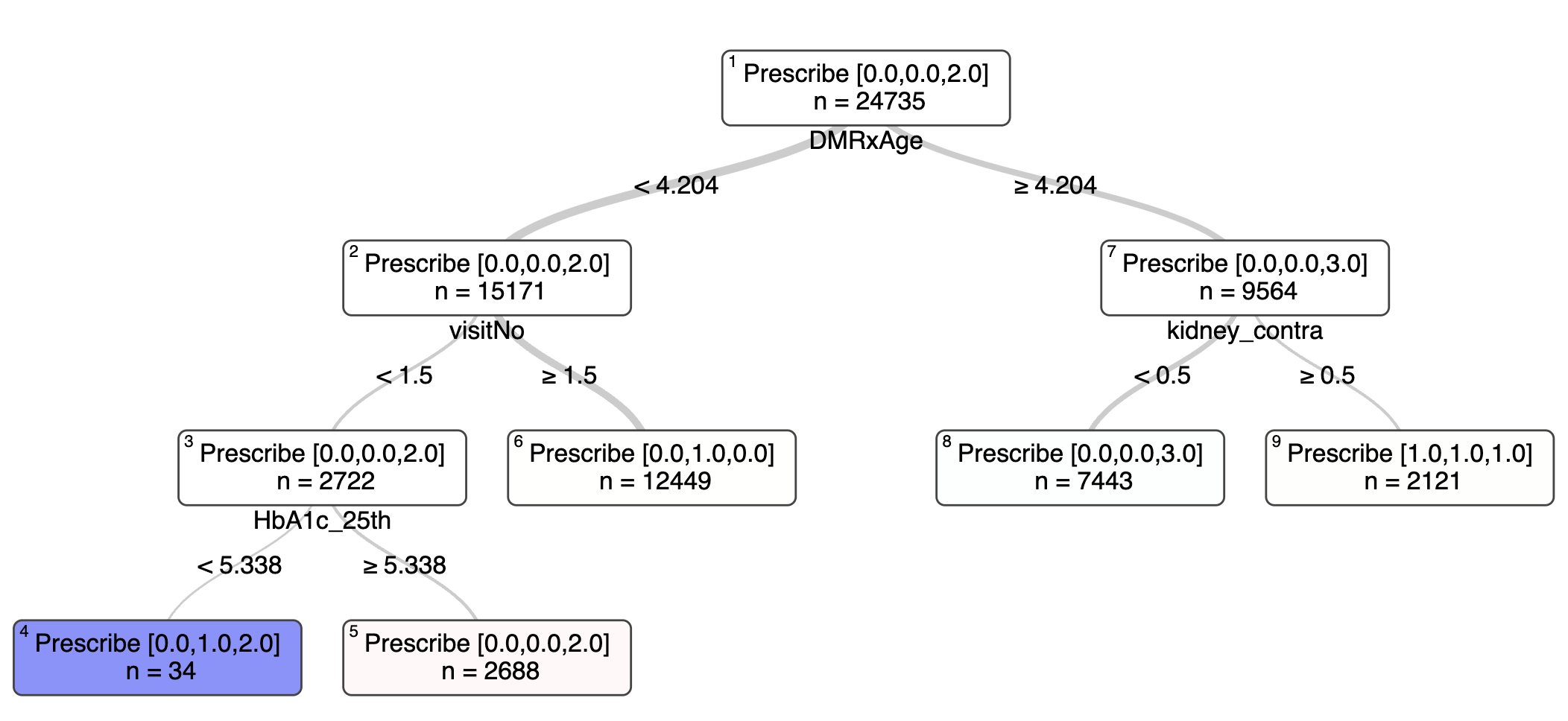}
  \end{center}
\end{sidewaysfigure}

\textcolor{white}{Force Appendix above figures}

\textcolor{white}{Force Appendix above figures}

\end{document}